\documentclass[acmlarge]{acmart}

\author{Shijie Chen}
\authornote{This work was done when the first author worked as a research assistant at Southern University of Science and Technology.}
\email{chen.10216@osu.edu}
\affiliation{
\institution{The Ohio State University}
\country{USA}
}

\author{Yu Zhang}
\authornote{Corresponding author}
\email{yu.zhang.ust@gmail.com}
\affiliation{
\institution{Southern University of Science and Technology}
\country{China}
}

\author{Qiang Yang}
\email{qyang@cse.ust.hk}
\affiliation{
\institution{Hong Kong University of Science and Technology}
\country{China}
}

\usepackage{times}
\usepackage{graphicx}
\usepackage{amsmath}
\usepackage{bm}
\usepackage{multirow}
\usepackage{amsthm}
\usepackage{url}
\usepackage{wrapfig}
\usepackage{multirow}
\usepackage{algorithm}
\usepackage{algpseudocode}
\usepackage{paralist}
\usepackage{hyperref}
\usepackage{subcaption}
\usepackage{xcolor}

\usepackage[final]{changes}

\makeatletter
\setdeletedmarkup{\@gobble{#1}}
\makeatother

\AtBeginDocument{\providecommand\BibTeX{{\normalfont B\kern-0.5em{\scshape i\kern-0.25em b}\kern-0.8em\TeX}}}
\newcommand{\vertiii}[1]{{\left\vert\kern-0.25ex\left\vert\kern-0.25ex\left\vert #1 \right\vert\kern-0.25ex\right\vert\kern-0.25ex\right\vert}}

\begin{document}

\begin{abstract}
Deep learning approaches have achieved great success in the field of Natural Language Processing (NLP). However, directly training deep neural models often suffer from overfitting and data scarcity problems that are pervasive in NLP tasks. In recent years, Multi-Task Learning (MTL), which can leverage useful information of related tasks to achieve simultaneous performance improvement on these tasks, has been used to handle these problems. In this paper, we give an overview of the use of MTL in NLP tasks. We first review MTL architectures used in NLP tasks and categorize them into four classes, including parallel architecture, hierarchical architecture, modular architecture, and generative adversarial architecture. Then we present optimization techniques on loss construction, gradient regularization, data sampling, and task scheduling to properly train a multi-task model. After presenting applications of MTL in a variety of NLP tasks, we introduce some benchmark datasets. Finally, we make a conclusion and discuss several possible research directions in this field.

%This paper provides an overview of the application of multi-task learning in natural language processing. In particular, we focus on mutli-task learning approaches based for deep learning models.
\end{abstract}
\date{}
\title{Multi-Task Learning in Natural Language Processing: An Overview}
\hypersetup{
 pdfauthor={Shijie Chen, Yu Zhang, Qiang Yang},
 pdftitle={Multi-Task Learning in Natural Language Processing: An Overview},
 pdflang={English}
 }

\maketitle

\section{Introduction}
\label{sec:org3c31405}

In recent years, data-driven neural models have achieved great success in machine learning problems.
In the field of Natural Language Processing (NLP), the introduction of transformers \cite{vspujgkp17} and pre-trained language models (PLMs) such as BERT \cite{dclt19}\added{, T5 \cite{2020t5} and GPT-3 \cite{NEURIPS2020_1457c0d6}} has led to a huge leap \replaced{in}{of} the performance \replaced{on}{in} multiple downstream tasks.
\added{While pre-training equips PLMs with general encyclopedic and linguistic knowledge, using PLMs on downstream tasks still requires task-specific adaptation.}
However, \added{sufficiently training} such models usually require a large amount of labeled training samples, which is often expensive for NLP tasks\deleted{ where linguistic knowledge is expected from annotators}.
\added{With the increasing size of neural models,} \replaced{t}{T}raining \replaced{them}{deep neural networks} on \replaced{downstream}{a large} dataset\added{s} also \replaced{demands}{asks for} immense computing power as well as huge time and storage budget.
To further improve\deleted{ the} model performance, combat the data scarcity problem, and facilitate cost-efficient \replaced{task adaptation}{machine learning}, researchers have adopted Multi-Task Learning (MTL) \cite{c97,zy21} for NLP tasks.
\added{More recently, with the uprising of generative pre-trained models \cite{2020t5,NEURIPS2020_1457c0d6}, notably large language models (LLMs), researchers have generalized the notion of performing tasks into following instructions \cite{mkbh22,xie-etal-2022-unifiedskg}, which virtually makes any NLP task a text-to-text task. This further allows to fine-tune a language model on a huge collection of tasks in a unified sequence-to-sequence framework. As a result, contemporary LLMs set new state-of-the-art on a variety of tasks and demonstrate an impressive ability in adapting to new tasks under few-shot and zero-shot settings  \cite{wei2022finetuned,sanh2022multitask}, highlighting the instrumental role of multi-task learning in building strong models for natural language processing.
}

MTL trains machine learning models from multiple related tasks simultaneously or enhances the model for a specific task using auxiliary tasks. Learning from multiple tasks makes it possible for models to capture generalized and complementary knowledge from the tasks at hand besides task-specific features.
Tasks in MTL can be tasks with assumed relatedness \cite{cw08,snrt15,gsa16,vvr17,lwwnw17}, tasks with different styles of supervision  (e.g., supervised and unsupervised tasks \cite{llsvk16,hzcyll16,llcp20}), tasks with different types of goals (e.g., classification and generation \cite{nmktmo19}), tasks with different levels of features (e.g., token-level and sentence-level features \cite{sg16,lgpe18}), and even tasks in different modalities (e.g., text and image data \cite{lysshzc16,skvbefl20}). Alternatively, we can treat the same task in multiple domains or languages as multiple tasks, which is also known as multi-domain learning \cite{yh15} in some literature, and learn an MTL model from them.

MTL naturally aggregates training samples from datasets of multiple tasks\added{ and alleviates the data scarcity problem}.
The benefit is escalated when unsupervised or self-supervised tasks, such as language modeling, are included. This is especially meaningful for low-resource tasks and languages whose labeled dataset is sometimes too small to sufficiently train a model.
In most cases, the enlarged training dataset \replaced{reduces}{alleviates} the risk of the overfitting and leads to more robust models. From this perspective, MTL acts similarly to data augmentation techniques \cite{gpb18}. However, MTL provides additional performance gain compared to data augmentation approaches, due to \replaced{its ability to learn common knowledge shared by different tasks}{the learned shared knowledge}.

While the thirst for better performance has driven people to build increasingly large models, developing more compact and efficient models with competitive performance has also received a growing interest\deleted{ in recent years}.
Through implicit knowledge sharing during the training process, MTL models could match or even exceed the performance of their single-task counterparts using much less training samples \cite{dh17,scn18}.
Besides, multi-task adapters \cite{sm19,pvgr20} transfer large pre-trained models to new tasks and languages by adding a modest amount of task-specific parameters. In this way, the costly fine-tuning of the entire model is avoided, which is important for real-world applications such as mobile computing and latency-sensitive services.
Many NLP models leverage additional features, including hand-crafted features and \replaced{those produced by}{output of} automatic NLP tools. Through MTL on \replaced{various}{different} linguistic tasks, such as chunking, Part-Of-Speech (POS) tagging, Named Entity Recognition (NER), and dependency parsing, we can reduce the reliance on external knowledge and prevent error propagation, which results in simpler models with potentially better performance \cite{lhoh18,zzw19,swr19,sxzcy20}.

This paper reviews the \replaced{application}{use} of MTL in recent NLP research. We focus on the ways in which researchers apply MTL to \added{downstream} NLP tasks, including model architecture\added{s}, training process\added{es}, and data source\added{s}. While most pre-trained language models take advantage of MTL during pre-training, they are not designed for specific down-stream tasks\added{,} and thus they are not in the focus of this paper. Depending on the objective of applying MTL, we denote by auxiliary MTL the case where auxiliary tasks are introduced to improve the performance of \deleted{one or more }primary tasks and by joint MTL the case where multiple tasks are equally important.

\deleted{In this paper, w}\added{W}e first introduce popular MTL architectures used in NLP tasks and categorize them into four classes, including parallel architecture, hierarchical architecture, modular architecture, and generative adversarial architecture \added{(Section \ref{sec:architecutres})}. Then we review optimization techniques of MTL for NLP tasks in terms of \deleted{the }loss construction, data sampling, and task scheduling \added{(Section \ref{sec:optimization})}. After that, we present applications of MTL, \replaced{categorized into auxiliary MTL and joint MTL}{which include auxiliary MTL and joint MTL as two main classes}, in a variety of NLP tasks \added{(Section \ref{sec:applications})}, and introduce some MTL benchmark datasets used in NLP \added{(Section \ref{sec:data})}. Finally, we conclude the whole paper and discuss several possible research topics in this field.

%The rest of this paper is organized as follows. Section \ref{sec:architecutres} presents the architecture of MTL models in NLP and Section \ref{sec:optimization} presents techniques for training MTL models. The applications of MTL in different NLP tasks are summarized in Section \ref{sec:applications}. At last, we introduce MTL benchmark datasets for the NLP field in Section \ref{sec:data} before making a conclusion in Section \ref{sec:conclusion}.

\noindent {\bf Notations}. In this paper, we use lowercase letters, such as $t$, to denote scal\replaced{a}{e}rs and use lowercase letters in boldface, such as $\mathbf{x}$, to denote vectors. Uppercase letters, such as $M$ and $T$, are used for constants and uppercase letters in boldface are used to represent matrices, including feature matrices like $\mathbf{X}$ and weight matrices like $\mathbf{W}$. In general, a multi-task learning model, parametrized by $\theta$, handles $M$ tasks on a dataset $\mathcal{D}$ with a loss function $\mathcal{L}$.

\section{MTL Architectures for NLP Tasks}
\label{sec:architecutres}

%Most research literature\deleted{s} focus\added{es} on the design of MTL architectures for NLP tasks. 

\added{The architectures of MTL models depend on the characteristics of the indented tasks as well as the design of the base models. When training generative models on instruction following, people usually train the entire model and focus more on data curation. We refer interested readers to another survey paper on instruction tuning \cite{zhang2023instruction}. In this work, we mainly focus on reviewing MTL architectures with task-specifc trainable parameters.
}

Based on how the relatedness between tasks are utilized, we categorize MTL architectures into the following classes: parallel architecture, hierarchical architecture, modular architecture, and generative adversarial architecture. 
The parallel architecture shares the bulk of the model among multiple tasks while each task has its own task-specific output layer.  
%[insights here?]
The hierarchical architecture models the hierarchical relationships between tasks. Such architecture can hierarchically combine features from different tasks, take the output of one task as the input of another task, or explicitly model the interaction between tasks.
%[insights here?]
The modular architecture decomposes the whole model into shared components and task-specific components that learn task-invariant and task-specific features, respectively.
%[insights here?]
Different from the above three architectures, the generative adversarial architecture borrows the idea of the generative adversarial network \cite{gpmxwocb14} to improve capabilities of existing models.
%[insights here?]
Note that the boundaries between different categories are not always solid and hence a specific model may fit into multiple classes. Still, we believe that this taxonomy could illustrate important ideas behind the design of MTL architectures.

Before introducing MTL architectures, we would like to clarify the definitions of hard and soft parameter sharing. In this paper, hard parameter sharing refers to sharing the same model parameters among multiple tasks, and it is the most widely used approach in multi-task learning models. Soft parameter sharing, on the other hand, constrains a distance metric between the intended parameters, such as the Euclidean distance \cite{gpb18b} and correlation matrix penalty \cite{hzcyll16}, to force certain parameters of models for different tasks to be similar. Alternatively, \added{\citet{ltn20}} add a regularization term to ensure the outputs of encoders of each task to be close for similar input instances. Differently, some researchers use hard parameter sharing to design a multi-task learning model that shares all the hidden layers except the final task-specific output layers and use soft parameter sharing to establish a multi-task model that partially shares \replaced{its parameters}{hidden layers} \cite{drls19}, such as embedding layers and low-level encoders. In this paper, such models fall into the `parallel architecture' category.

\subsection{Parallel Architectures}
\label{sec:orgd2c2890}
As its name suggests, the model for each task run in parallel under the parallel architecture, which is implemented by sharing certain intermediate layers. In this case, there is no dependency other than layer sharing among tasks. Therefore, there is no constraint on the order of training samples from each task. During training, the shared parameters receive gradients from samples of each task, enabling knowledge sharing among tasks. Fig. \ref{fig:parallel_arch} illustrates different forms of parallel architectures.

\subsubsection{Parallel Feature Sharing.}
\label{sec:org5bd297b}
The simplest form of parallel architecture is a parallel feature sharing architecture (Fig. \ref{fig:treelike}), where the models for different tasks share a base feature extractor (i.e., the trunk) followed by task-specific encoders and output layers (i.e., the branches). A shallow trunk can be simply the word representation layer \cite{scn18} while a deep trunk can be the entire model except output layers. The tree-like architecture was proposed by \added{\citet{c97}} and has been
%widely used in MTL %\cite{wh15,llsvk16,bs16,gsa16,czb16,lqh16,as17,vvr17,hxts17,mthma18,td18,fov18,lhoh18,cjol18,klz18,gpb18b,lwscdl18,zfszwy18,ftl19,rly19,nnnosat19,zlzw19,slf19,ylxscz19,zh19,nmktmo19,sgqgtdlj19,wtnmt19,cbyls20,zhzlx20,jgkch20,sxzcy20,clthhz20,creb20,wclqlw20,wzlsz20,llzs16,pts17,wfgwz18,zcq18,ks19}.
widely used in MTL \cite{lz08,wh15,llsvk16,bs16,gsa16,czb16,lqh16,as17,vvr17,hxts17,mthma18,td18,fov18,lhoh18,cjol18,klz18,gpb18b,lwscdl18,zfszwy18,ftl19,rly19,nnnosat19,zlzw19,pb19,slf19,ylxscz19,zh19,nmktmo19,sgqgtdlj19,wtnmt19,cbyls20,zhzlx20,jgkch20,sxzcy20,clthhz20,creb20,wclqlw20,wzlsz20,llzs16,pts17,wfgwz18,zcq18,ks19,ard20}.
In some literature, this architecture is also known as hard sharing architecture or multi-head architecture, where each head corresponds to the combination of a task-specific encoder and the corresponding output layer or just a branch.

Parallel feature sharing uses a single trunk to force all tasks to share the same low-level feature representation, which may limit the expressive power of the model for each task. A solution is to equip the shared trunk with task-specific encoders \cite{xzz18,har18,ltn20}. For example, \added{\citet{lysj18}} combine a shared character embedding layer and language-specific word embedding layers for different languages.
Another way is to make different groups of tasks share different parts of the trunk \cite{pb17,gpb18,mthma18}. \replaced{T}{Moreover, t}his idea can \added{also} be applied to the decoder. For instance, \added{\citet{wza20}} share the trunk encoder with a source-side language model and shares the decoder with a target-side denoising autoencoder.

\subsubsection{Parallel Feature Fusion}
\label{sec:org67a2823}
\begin{figure}
     \centering
     \begin{subfigure}[t]{.33\linewidth}
         \centering
         \includegraphics[width=\linewidth]{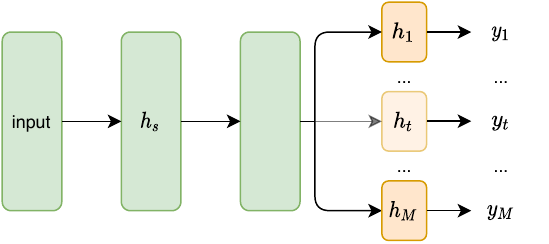}
         \caption{Parallel Feature Sharing}
         \label{fig:treelike}
     \end{subfigure}%
     \begin{subfigure}[t]{.33\linewidth}
         \centering
         \includegraphics[width=\linewidth]{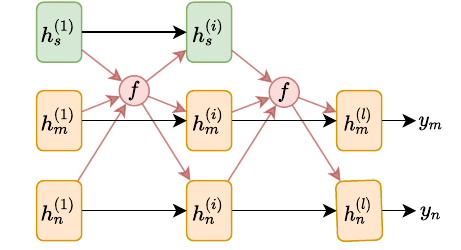}
         \caption{Parallel Feature Fusion}
         \label{fig:paraFF}
     \end{subfigure}
     \begin{subfigure}[t]{.33\linewidth}
         \centering
         \includegraphics[width=\linewidth]{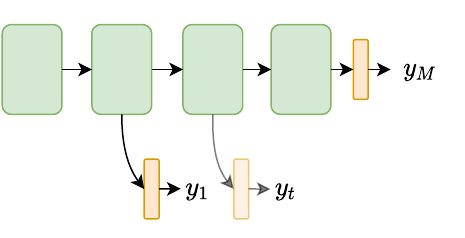}
         \caption{Parallel Multi-level Supervision}
         \label{fig:supDiffLevels}
     \end{subfigure}
        \caption{Illustration for parallel architectures. For task $t$, $h_t^{(i)}$ represents the latent representation at the $i$-th layer and $y_t$ represents the corresponding label ($h_s$ are shared latent representations). 
        The green blocks represent shared parameters and the orange blocks are task-specific parameters. Red circles represent feature fusion mechanism $f$.}
        \label{fig:parallel_arch}
\end{figure}

Different from learning shared features implicitly by sharing model parameters in the trunk, MTL models can actively combine features from different tasks, including shared and task-specific features, to form representations for each task. As shown in Fig. \ref{fig:paraFF}, such models can use a globally shared encoder to produce shared representations that can be used as additional features for each task-specific model \cite{lqh16}. The shared representations can also be used indirectly as the key for attention layers in each task-specific model \cite{tzwx19}. 

However, simply aggregating features of different tasks via weighted sum \cite{ll17} or attention \cite{zcq18} is sub-optimal since these features might actually hurt the performance of other tasks, \replaced{also}{a phenomena} known as inter-task interference.
Researchers have proposed \replaced{to do}{several ways to alleviate the inter-task interference via} more fine-grained feature sharing between tasks\added{ to counter this issue}.
One approach is to directly \replaced{aggregate}{aggregation} shared and task-specific features using learnable feed-forward layers \cite{zxwj17,gcc19} or gating mechanisms \cite{lwwnw17,drls19}. Additionally, feature sharing can \deleted{also }be indirectly performed by maintaining memory units that are shared among different tasks either globally or \replaced{in pairs}{pairwise} \cite{lqh16,wrjns19}.

A more generalized approach for inter-task feature sharing is modeling task relatedness and sharing features accordingly. As an example, Sluice network \cite{rbas19} controls feature transfer by a learned task relatedness matrix.
Instead of using a fixed relatedness matrix, LK-MTL \cite{xzcwj18} uses leaky units to dynamically control pairwise feature flow based on input features, and similar to RNN cells, it modulates information flow by two gates. Specifically, given two tasks \(m\) and \(n\), the leaky gate \(\mathbf{r}_{mn}\) determines how much knowledge should be transferred from task \(n\) to task \(m\) and \deleted{then it }emits a feature map \(\tilde{\mathbf{h}}_{mn}\). The update gate \(\mathbf{z}_{mn}\) determines how much information should be maintained from task \(m\) and \deleted{then it }emits \added{the }final output \(\tilde{\mathbf{h}}_m\) for task \(m\). Mathematically, the \replaced{feature sharing process is }{two gates can be }formulated as\added{:}
\begin{equation*}
\begin{aligned}
\mathbf{r}_{mn} &= \sigma(\mathbf{W}_r\cdot[\mathbf{h}_m,\mathbf{h}_n])\\
\tilde{\mathbf{h}}_{mn} &= \mathrm{tanh}(\mathbf{U} \cdot \mathbf{h}_m + \mathbf{W} \cdot (\mathbf{r}_{mn}\odot \mathbf{h}_n))\\
\mathbf{z}_{mn} &= \sigma(\mathbf{W}_z\cdot[\mathbf{h}_m,\mathbf{h}_n])\\
\tilde{\mathbf{h}}_m &= \mathbf{z}_{mn}\cdot \mathbf{h}_m + (1-\mathbf{z}_{mn})\cdot\tilde{\mathbf{h}}_{mn},
\end{aligned}
\end{equation*}
where $\sigma(\cdot)$ denotes the sigmoid function and $\mathrm{tanh}(\cdot)$ denotes the hyperbolic tangent function.
When considering all pairwise directions, the output for each task is given by the sum of each row in
\begin{equation*}
\left[\begin{array}{cccc}
\sum_{k=1}^{M} \mathbf{z}_{1 k} & \left(1-\mathbf{z}_{12}\right) & \cdots & \left(1-\mathbf{z}_{1 M}\right) \\
\left(1-\mathbf{z}_{21}\right) & \sum_{k=1}^{M} \mathbf{z}_{2 k} & \cdot & \left.1-\mathbf{z}_{2 M}\right) \\
\vdots & \vdots & \ddots & \vdots \\
\left(1-\mathbf{z}_{M 1}\right) & \left(1-\mathbf{z}_{M 2}\right) & \cdots & \sum_{k=1}^{M} \mathbf{z}_{M k}
\end{array}\right] \cdot\left[\begin{array}{cccc}
\mathbf{h}_{1} & \mathbf{h}_{12} & \cdots & \mathbf{h}_{1 M} \\
\tilde{\mathbf{h}}_{21}^{i} & \mathbf{h}_{2} & \cdots & \tilde{\mathbf{h}}_{2 M}^{i} \\
\vdots & \vdots & \ddots & \vdots \\
\tilde{\mathbf{h}}_{M 1}^{i} & \tilde{\mathbf{h}}_{M 2}^{i} & \cdots & \tilde{\mathbf{h}}_{M}^{i}
\end{array}\right] / M.
\end{equation*}

Task routing is another method for dynamic feature fusion, where the paths that samples go through in the model differ by their task\added{s}.
Given \(M\) tasks, the routing network \added{in} \cite{zbh18} splits RNN cells into several shared blocks with \(M\) task-specific blocks (one for each task) and then modulates the input to as well as output from each RNN block by a learned weight.
MCapsNet \cite{xzcwj18b}, which adapts CapsNet \cite{sfh17} to NLP tasks, replaces dynamic routing in CapsNet with task routing to build different feature spaces for each task. In MCapsNet, similar to dynamic routing, task routing computes task coupling coefficients \(c_{ij}^{(k)}\) for capsule \(i\) in the current layer and capsule \(j\) in the next layer for task \(k\). 
Due to the fine-grained dynamic control of information flow between tasks, LK-MTL and MCapsNet outperform other feature fusion methods and obtain state-of-the-art performance.

%\subsubsection{Supervision at Different Feature Levels}
\subsubsection{Parallel Multi-level Supervision.}
\label{sec:orga175ff6}
\replaced{While m}{M}odels using the parallel architecture handle multiple tasks in parallel\replaced{, t}{. T}hese tasks may concern features at different abstraction levels. For NLP tasks, such levels can be character-level, token-level, sentence-level, paragraph-level, and document-level. Due to the compositional nature of language\added{,} both syntactically and semantically, it is natural to give supervision signals at different depths of an MTL model for tasks at different levels \cite{cw08,sg16,mtdnd18,swr19} as illustrated in Fig. \ref{fig:supDiffLevels}.
For example, in \cite{lgpe18,fy19}, token-level tasks receive supervisions at lower-layers while sentence-level tasks receive supervision at higher layers. 
\added{\citet{rly19}} supervise a higher-level QA task on both sentence and document-level features in addition to a sentence similarity prediction task that only relies on sentence-level features.
In addition, \added{\citet{glzolzzdc19,pcks18}} add skip connections so that signals from higher-level tasks are amplified.
\added{\citet{clspb20}} learn \deleted{the task of}semantic goal navigation at a lower level and learns the task of embodied question answering at a higher level.

In some settings where MTL is used to improve the performance of a primary task, the introduction of auxiliary tasks at different levels could be helpful. Several works integrate a language modeling task on lower-level encoders for better performance on  simile detection \cite{rei17}, sequence labeling \cite{lhsflh18}, question generation \cite{zzw19}, and task-oriented dialogue generation \cite{zzw19}. \added{\citet{lc19}} add sentence-level sentiment classification and attention-level supervision to assist the primary stance detection task. \added{\citet{nmktmo19}} add attention-level supervision to improve consistency of the two primary language generation tasks. \added{\citet{csll20}} minimize an auxiliary cosine softmax loss based on the audio encoder to learn more accurate speech-to-semantic mappings.

\subsection{Hierarchical Architectures}
\label{sec:orgeb86fed}

\begin{figure}
     \centering
     \begin{subfigure}[t]{.33\linewidth}
         \centering
         \includegraphics[scale=0.65]{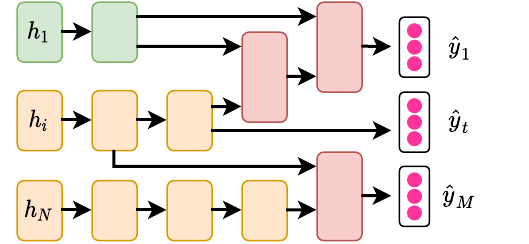}
         \caption{Hierarchical Feature Fusion}
     \end{subfigure}%
     \begin{subfigure}[t]{.33\linewidth}
         \centering
         \includegraphics[scale=.65]{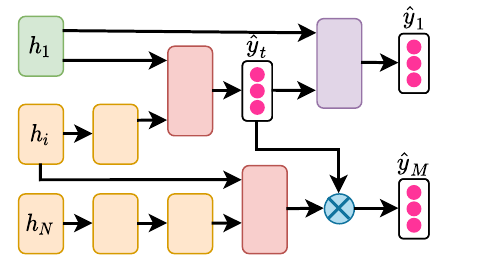}
         \caption{Hierarchical Pipeline}
     \end{subfigure}
     \begin{subfigure}[t]{.33\linewidth}
         \centering
         \includegraphics[scale=.65]{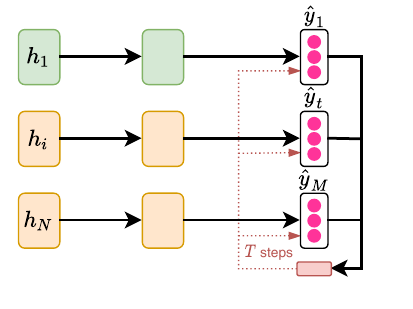}
         \caption{Hierarchical Interactive MTL}
     \end{subfigure}
        \caption{Illustration for hierarchical architectures. \(h\) represents different hidden states and \(\hat y_t\) represents the predicted output distribution for task \(t\). Red boxes stand for hierarchical feature fusion mechanisms. The purple block and blue circle in (b) stand for hierarchical feature and signal pipeline unit respectively.}
        \label{fig:hier_arch}
\end{figure}

The hierarchical architecture considers hierarchical relationships among multiple tasks. The features and output of one task can be used by another task as an extra input or additional control signals. The design of hierarchical architectures depends on the tasks at hand and is usually more complicated than parallel architectures. Fig. \ref{fig:hier_arch} illustrates different hierarchical architectures. 
We notice that parallel MTL architectures usually assume the features shared are in the same feature space. Thus they should be processed by similar model architectures. In contrast, Hierarchical MTL architectures allow independent processing for each task and could accommodate tasks with data in heterogeneous feature spaces such as text, knowledge graphs, images, and audio. 

\subsubsection{Hierarchical Feature Fusion}
\label{sec:org8e90193}

Different from parallel feature fusion that combines features of different tasks at the same depth, hierarchical feature fusion can explicitly combine features at different depths and allow different processing for different features. To solve the Twitter demographic classification problem, \added{\citet{vvr17}} encode the name, following network, profile description, and profile picture features of each user by different neural models and combines the outputs using an attention mechanism.
\added{\citet{lhsflh18}} take the hidden states for tokens in simile extraction as an extra feature in the sentence-level simile classification task.
For knowledge base question answering, \added{\citet{dxlydfls19}} combine lower level word and knowledge features with more abstract semantic and knowledge semantic features by a weighted sum.
\added{\cite{wwsllszz20}} fuses topic features of different roles into the main model via a gating mechanism.
In \cite{creb20}, text and video features are combined through inter-modal attention mechanisms of different granularity to improve performance of sarcasm detection.

\subsubsection{Hierarchical Pipeline}
\label{sec:orgdd25c2c}

Instead of aggregating features from different tasks as in feature fusion architectures, pipeline architectures treat the output of a task as an extra input of another task and form a hierarchical pipeline between tasks. In this section, we refer to \emph{output} as the final result \replaced{for}{from} a task, including the final output distribution and hidden states before the last output layer. \replaced{The extra input can be used directly as input features or used indirectly as control signals to enhance the performance of other tasks. Therefore, w}{W}e further divide hierarchical pipeline architectures into hierarchical feature pipeline and hierarchical signal pipeline.

In hierarchical feature pipeline, the output of one task is used as extra features for another task. The tasks are assumed to be directly related so that outputs instead of hidden feature representations are helpful to other tasks.
For example, \added{\citet{cgzzwzs19}} feed the output of a question-review pair recognition model to the question answering model. %In many cases, the tasks are at the same abstraction level.
\added{\citet{hlnd19}} feed the output of aspect term extraction to aspect-term sentiment classification.
Targeting community question answering, \added{\citet{yccwzs19}} use the result of question category prediction to enhance document representations.
\added{\citet{sp19}} feed the result of morphological tagging to a POS tagging model and the two models are further tied by skip connections. 
%To enable asynchronous training of multi-task models, \cite{zlzw19} initializes input from other tasks by a uniform distribution across labels.

Hierarchical feature pipeline is especially useful for tasks at different abstraction levels. \added{\citet{ftl19}} use the output of neighboring word semantic type prediction as extra features for neighboring word prediction.
\added{\citet{hxts17}} use skip connections to forward predictions of lower-level POS tagging, chunking, and dependency parsing tasks to higher-level entailment and relatedness classification tasks.
In addition, deep cascade MTL \cite{glzolzzdc19} adds both residual connections and cascade connections to a single-trunk parallel MTL model with supervision at different levels, where residual connections forward hidden representations and cascade connections forward output distributions of a task to the prediction layer of another task.
\added{\citet{sslf20}} include the output of the low-level discourse element identification task in the organization grid, which consists of sentence-level, phrase-level, and document-level features of an essay, for the primary essay organization evaluation task.
In \cite{slf19}, the word predominant sense prediction task and the text categorization task share a transformer-based embedding layer and embeddings of certain words in the text categorization task could be replaced by prediction results of the predominant sense prediction task.

The direction of hierarchical pipelines is not necessarily always from low-level tasks to high-level tasks. For example, in \cite{amd20}, the outputs of word-level tasks are fed to the char-level primary task.
\added{\citet{rboc20}} feed the output of more general classification models to more specific classification models during training, and the more general classification results are used to optimize beam search of more specific models at test time.

In hierarchical signal pipeline, the outputs of tasks are used indirectly as external signals to help improve the performance of other tasks.
For example, the predicted probability of the sentence extraction task \replaced{can be}{is} used to weigh sentence embeddings for a document-level classification task \cite{ifmms17}.
For the hashtag segmentation task, \added{\citet{mxp19}} first predict the probability of a hashtag being single-token or multi-token as an auxiliary task and further use the output to combine single-token and multi-token features.
In \cite{sgqgtdlj19}, the output of an auxiliary entity type prediction task is used to disambiguate candidate entities for logical form prediction.
The outputs of a task can also be used for post-processing. For instance, \added{\citet{zzl20}} use the output of NER to help extract multi-token entities.

\subsubsection{Hierarchical Interactive MTL}
\label{sec:org032e334}

Different from most machine learning models that give predictions in a single pass, hierarchical interactive MTL explicitly models the interactions between tasks via a multi-turn prediction mechanism which allows a model to refine its predictions over multiple steps with the help of the previous outputs from other tasks in a way similar to recurrent neural networks.
\added{\citet{hlnd19}} maintain a shared latent representation which is updated by \(T\) iterations.
In cyclic MTL \cite{zssxsl20}, the output of one task is used as an extra input to its successive lower-level task and the output of the last task is fed to the first one, forming a loop.
Most hierarchical interactive MTL models as introduced above report that performance converges quickly at \(T=2\) steps, showing the benefit and efficiency of doing multi-step prediction.

\subsection{Modular Architectures}
\label{sec:org4ea85d9}

The idea behind the modular MTL architecture is simple: breaking an MTL model into shared modules and task-specific modules. The shared modules learn shared features from multiple tasks. Since the shared modules can learn from many tasks, they can be sufficiently trained and can generalize better, which is particularly meaningful for low-resource scenarios. On the other hand, task-specific modules learn features that are specific to a certain task. Compared with shared modules, task-specific modules are usually much smaller and thus less likely to suffer from overfitting caused by insufficient training data. The robustness of shared modules and the flexibility of task-specific modules makes modular architectures suitable for learning different tasks efficiently. 
%Here each module can have various structures depending on different designs, from an embedding layer to an entire block of several layers. %For instance, \cite{wh16} assigns a separate classifier for each user in addition to a globally shared classifier to cover the diversity of Microblog users, and due to the large amount of user-specific classifiers, a distributed training approach is applied to improve the training efficiency.

The simplest form of modular architectures is a single shared module coupled with task-specific modules as in parallel feature sharing described in Section \ref{sec:org5bd297b}.
Besides, another common practice is to share the first embedding layers across tasks \deleted{as }\cite{zl19,ltn20}\deleted{ did}.
\added{\citet{amd20}} share word and character embedding matrices and combines them differently for different tasks.
\added{\citet{sba19}} share two encoding layers and a vocabulary lookup table between the primary neural machine translation task and the auxiliary representation learning task. 
Shared embeddings can be used alongside task-specific embeddings \cite{lzs19,yesb19} as well. In addition to word embeddings, \cite{zxcwj18} shares label embeddings between tasks.
Researchers have also developed modular architectures at a finer granularity.
For example, \added{\citet{tfszy18}} split the model into task-specific encoders and language-specific encoders for \replaced{multilingual}{multi-lingual} dialogue evaluation. In \cite{dxlydfls19}, each task has its own encoder and decoder, while all tasks share a representation learning layer and a joint encoding layer. 
\added{\citet{pld19}} create encoder modules on different levels, including task level, task group level, and universal level.

\begin{figure}
     \centering
     \begin{subfigure}[t]{.49\linewidth}
         \centering
         \includegraphics[scale=.5]{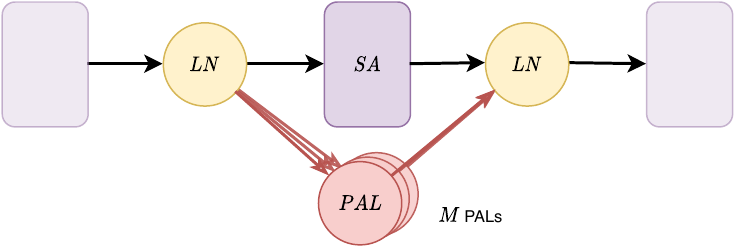}
         \caption{Bert and PALs \cite{sm19}}
     \end{subfigure}%
     \begin{subfigure}[t]{.49\linewidth}
         \centering
         \includegraphics[scale=.5]{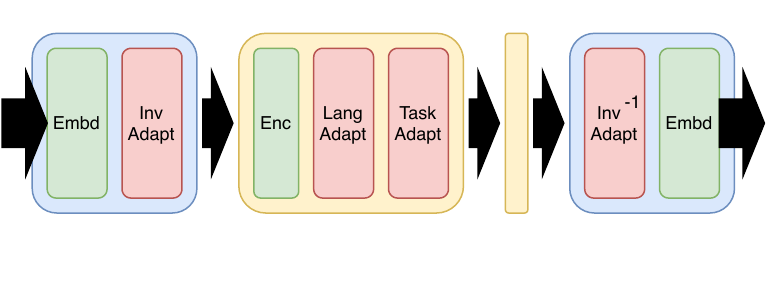}
         \caption{MAD-X \cite{pvgr20}}
     \end{subfigure}
     \caption{Illustration for multi-task adapters.}
     \label{fig:modular_arch}
\end{figure}

When adapting large pre-trained models to down-stream tasks, a common practice is to fine-tune a separate model for each task. While this approach usually attains good performance, it poses heavy computational and storage costs. A more cost-efficient way is to add \added{lightweight} task-specific \added{trainable} modules into a single shared \added{frozen} \replaced{backbone}{pre-trained} model.
\added{A special case is prefix-tuning for adapting pre-trained generative language models \cite{li-liang-2021-prefix}, where learnable prefix vectors are prepended to inputs to frozen language models as context. Several works train task-specific prompt vectors for MTL \cite{vu-etal-2022-spot,asai-etal-2022-attempt}. \citet{wang2023multitask} further improve multi-task prefix-tuning by decomposing the task prompts into a task-shared prompt and smaller task-specific prompts. }

\deleted{As an example,}\replaced{M}{m}ulti-task adapters adapt single-task models to multiple tasks by adding extra task-specific parameters (adapters).
\added{\citet{sm19}} add task-specific Projected Attention Layers (PALs) in parallel with self-attention operations in a pre-trained BERT model. Here PALs in different layers share the same parameters to reduce model capacity and improve training speed.
In Multiple ADapters for Cross-lingual transfer (MAD-X) \cite{pvgr20}, the model is decomposed into four types of adapters: language adapters, task adapters, invertible adapters, and its counterpart inversed adapters, where language adapters learn language-specific task-invariant features, task adapters learn language-invariant task-specific features, invertible adapters conversely map input embeddings from different tasks into a shared feature space, and inversed adapters map hidden states into domain-specific embeddings.
MAD-X can perform quick domain adaptation by directly switching corresponding language and task adapters instead of training new models from the scratch.

Further, task adaptation modules can also be dynamically generated by a meta-network. As an example, 
\replaced{H}{h}ypergrid transformer \cite{tzbmj20} scales the weight matrix $H$ of the second feed forward layer in each transformer block by the multiplication of two vectors as
\begin{equation*}
  \mathbf{H}(\mathbf{x}) = \phi(\sigma((\mathbf{L}_{row}\cdot\mathbf{x})(\mathbf{L}_{col}\cdot\mathbf{x})))\odot \mathbf{W},
\end{equation*}
where $\mathbf{L}_{row}$ and $\mathbf{L}_{col}$ are either globally shared task feature vectors or local instance-wise feature vectors, $\phi$ is a scaling operation, $\mathbf{x}$ is an input vector, and $\mathbf{W}$ is a learnable weight matrix.
\added{Similarly, Hyperformer \cite{karimi2021parameterefficient} inserts feed-forward adapter modules, which are generated by a task-aware hypernetwork, between pre-trained Transformer layers for efficient adaptation.}
Differently, Conditionally Adaptive MTL (CA-MTL) \cite{php21} implements task adapters in the self-attention operation of each transformer block based on task representations $\{\mathbf{z}_{i}\}$ as
\begin{equation*}
\mathrm{Attention}\left(\mathbf{Q},\mathbf{K},\mathbf{V},\mathbf{z}_{i}\right)=\mathrm{softmax}\left(\mathbf{M}\left(\mathbf{z}_{i}\right)+\frac{\mathbf{Q} \mathbf{K}^{T}}{\sqrt{d}}\right)\mathbf{V}
\end{equation*}
where $\mathbf{M}(\mathbf{z}_{i}) = \mathrm{diag}(\mathbf{A}^\prime_{1}(\mathbf{z}_{i}),\dots,\mathbf{A}^{\prime}_{N}(\mathbf{z}_{i}))$ is a diagonal block matrix consisting of $N$ learnable linear transformations over $\mathbf{z}_{i}$. Therefore, $\mathbf{M}(\mathbf{z}_{i})$ injects task-specific bias into the attention map in the self-attention mechanism. Similar adaptation operations are used in input alignment and layer normalization as well.
Impressively, a single jointly trained Hypergrid transformer, \added{Hyperformer,} or CA-MTL model could match or outperform single-task fine-tuned models on multi-task benchmark datasets while only adding a negligible amount of parameters.
\added{Instead of generating adaptation parameters with hypernetworks, Mixture-of-Expert (MoE) models \cite{shazeer2017} adjust computation by routing input to different trainable expert modules and show performance improvement on MTL \cite{kim2021scalable,gao-etal-2022-parameter,zhao2023jiuzhang}. More recently, task-specific information has been introduced to the routing algorithm for further performance improvement \cite{gupta2022sparsely,pham-etal-2023-task}.}

\subsection{Generative Adversarial Architectures}
\label{sec:orgef93167}

\deleted{Recently }Generative Adversarial Networks (GANs) have achieved great success in generative tasks for computer vision. The basic idea of GANs is to train a discriminator \added{model} that distinguishes generated images from ground truth ones\added{ and train the generator model to fool the discriminator}. By \replaced{jointly optimizing both models}{the discriminator}, we can obtain a generator that can produce more vivid images and a discriminator that is better at spotting synthesized images.
A similar idea can be used in MTL for NLP tasks. By introducing a discriminator \(G\) that predicts which task a given training instance comes from, the shared feature extractor \(E\) is forced to produce more generalized task-invariant features \cite{lqh17,wfgwz18,msha18,tfszy18,yesb19} and therefore improve the performance and robustness of the entire \added{MTL} model.
In the training process of such models, the adversarial objective is usually formulated as
\begin{equation*}\mathcal{L}_{adv} =\min_{\theta_E}\max_{\theta_D} \sum_{t=1}^{M}\sum_{i=1}^{|\mathcal{D}_t|}d_i^t\log[D(E(\mathbf{X}))],
\end{equation*}
where \(\theta_E\) and \(\theta_D\) denote model parameters for the feature extractor and discriminator, respectively, and \(d_i^t\) denotes the one-hot task label.

An additional benefit of generative adversarial architectures is that unlabeled data can be fully utilized.
\added{\citet{wwsllszz20}} add an auxiliary generative model that reconstructs documents from document representations learned by the primary model and improves the quality of document representations by training the generative model on unlabeled documents.
To improve the performance of an extractive machine reading comprehension model, \added{\citet{rcs20}} use a self-supervised approach. First, a discriminator that rates the quality of candidate answers is trained on labeled samples. Then, during unsupervised adversarial training, the answer extractor tries to obtain a high score from the discriminator.

\section{Optimization for MTL Models}
\label{sec:optimization}

Optimization techniques of training MTL models are equally as important as the design of model architectures. In this section, we summarize optimization techniques for MTL models used in recent research literatures targeting NLP tasks, including loss construction, data sampling, and task scheduling.

\subsection{Loss Construction}
\label{sec:org0282301}

The most common approach to train an MTL model is to linearly combine loss functions of different tasks into a single global loss function. In this way, the entire objective function of the MTL model can be optimized through conventional learning techniques such as stochastic gradient descent with back-propagation.
Different tasks may use different types of loss functions. For example, in \cite{ylxscz19}, the cross-entropy loss for the relation identification task and the ranking loss for the relation classification task are linearly combined, which performs better than single-task learning. Specifically, given \(M\) tasks each associated with a loss function \(\mathcal{L}_i\) and a weight \(\lambda_t\), the overall loss \(\mathcal{L}\) is defined as
\begin{equation*}
\mathcal{L} = \sum_{t=1}^M\lambda_t\mathcal{L}_t + \sum\lambda_a\mathcal{L}_{adap} + \sum\lambda_r\mathcal{L}_{reg},
\end{equation*}
where \(\mathcal{L}_t\), \(\mathcal{L}_{adap}\), and \(\mathcal{L}_{r}\) denotes loss functions of different tasks, adaptive losses, and regularization terms, \deleted{respectively, }with \(\lambda_t\), \(\lambda_a\), and \(\lambda_{reg}\) being their respective weights. For cases where the tasks are optimized in turns \added{rather than joint training \cite{stbp18}}\deleted{as in \cite{stbp18} instead of being optimized jointly}, \(\lambda_t\) is equivalent to the sampling weight $p_t$ for task \(t\), which will be discussed in Section \ref{sec:orge551827}.

\replaced{A}{In this case, a}n important question is how to assign a proper weight \(\lambda_t\) to each task. The simplest way is to set them equally \cite{pts17,zl19,wzlsz20}, i.e., \(\lambda_t=\frac{1}{M}\).
As a generalization, the weights are usually viewed as hyper-parameters and set based on experience or through grid search \cite{lqh16,lwwnw17,lwscdl18,zxcwj18,cqlh18,zfszwy18,lhoh18,lhsflh18,sgbjcld19,gcc19,mxp19,nnnosat19,sba19,yesb19,fy19,drls19,zzw19,zzh19,sgqgtdlj19,xlz19,dxlydfls19,wrjns19,zssxsl20,zssxsl20,zzl20,wwsllszz20,rcs20,clthhz20,cbyls20,zhzlx20,sxzcy20}.
For example, to prevent large datasets from dominating training, \added{\citet{pcks18}} set the weights as
\begin{equation*}
\lambda_t \propto \frac{1}{|\mathcal{D}_t|}\ ,
\end{equation*}
where \(|\mathcal{D}_t|\) denotes the size of the training dataset for task \(t\).
The weights can also be adjusted dynamically during the training process based on certain metrics\replaced{. T}{and t}hrough adjusting weights, we can purposely emphasize different tasks in different training stages.
For instance, since dynamically assigning smaller weights to more uncertain tasks usually leads to good performance for MTL \deleted{according to }\cite{cgk18}, \cite{lgpe18} assigns weights based on the homoscedasticity of training losses from different tasks as
\begin{equation*}
\lambda_t=\frac{1}{2\sigma_t^2},
\end{equation*}
where \(\sigma_t\) measures the variance of the training loss for task \(t\).
In \cite{llcp20}, the weight of an unsupervised task is set to a confidence score that measures how much a prediction resembles the corresponding self-supervised label.
% that is, 
% \begin{equation*}
% \lambda = g(\hat{\mathbf{y}}^*,\hat{\mathbf{y}}),
% \end{equation*}
%which assumes that the model can more confidently learn from unlabeled samples if the prediction has a similar structure to the extracted label.
To ensure that a student model could receive enough supervision during knowledge distillation, BAM! \cite{clkml19} combines the supervised loss \(\mathcal{L}_{sup}\) with the distillation loss \(\mathcal{L}_{diss}\) as
\begin{equation*}
\mathcal{L} = \lambda\mathcal{L}_{diss} + (1-\lambda)\mathcal{L} _{sup},
\end{equation*}
where \(\lambda\) increases linearly from 0 to 1 in the training process.
In \cite{sslf20}, three tasks are jointly optimized, including the primary essay organization evaluation (OE) task \replaced{as well as}{and} the auxiliary sentence function identification (SFI) and paragraph function identification (PFI) tasks. The two lower-level auxiliary tasks are assumed to be equally important with weights set to 1 (i.e., $\lambda_{SFI}=\lambda_{PFI}=1$)  and the weight of the OE task is set as
\begin{equation*}
\lambda_{OE}=\max \left(\min \left(\frac{\mathcal{L}_{O E}}{\mathcal{L}_{S F I}} \cdot \lambda_{OE}, 1\right), 0.01\right),
\end{equation*}
where \(\lambda_{OE}\) is initialized to 0.1 and then dynamically updated \added{during training,} so that the model focuses on the lower-level tasks at first before \(\lambda_{OE}\) becomes larger when \(\mathcal{L}_{SFI}\) gets relatively smaller. \added{\citet{nmktmo19}} guide the model to focus on easy tasks by setting weights as
\begin{equation}\lambda_t(e) = \frac{\lambda_t^{const}}{1+\exp((e_t'-e)/\alpha)},
\label{sigmoid_scheduling}
\end{equation}
where \(e\) denotes the number of epochs,  \(\lambda_t^{const}\) and \(e_t'\) are hyperparameters for each task, and \(\alpha\) denotes temperature.

In addition to combining loss functions from different tasks, researchers also use additional adaptive loss functions \(\mathcal{L}_{adapt}\) to enhance MTL models.
In \cite{lc19}, the alignment between an attention vector and a hand-crafted lexicon feature vector is normalized to encourage the model to attend to important words in the input.
\added{\citet{cgzzwzs19}} penalize the similarity between attention vectors from two tasks and the Euclidean distance between the resulting feature representations to enforce the models to focus on different task-specific features.
To learn domain-invariant features, \added{\citet{xzz18}} minimize a distance function \(g(\cdot)\)  between a pair of learned representations from different tasks. Candidates of \(g(\cdot)\) include the KL divergence, maximum mean discrepancy (MMD), and central moment discrepancy (CMD). Extensive experiments \replaced{show}{report} that KL divergence gives overall stable improvements on all experiments while CMD hits more best scores.

The \(L_1\) metric linearly combines different loss functions and optimizes all tasks simultaneously. However, when we view multi-task learning as a multi-objective optimization problem, this type of objective functions cannot guarantee optimality in obtaining Pareto-optimal models when each loss function is non-convex. To address this issue, Tchebycheff loss \cite{myld20} optimizes an MTL model by an  \(L_\infty\) objective,  which is formulated as \begin{equation*}
{\mathcal{L}}_{c h e b}=\max _{t}\left\{\lambda_{1} {\mathcal{L}}_{1}\left(\theta^{s h}, \theta^{1}\right), \ldots, \lambda_{M} {\mathcal{L}}_{M}\left(\theta^{s h}, \theta^{M}\right)\right\}
\end{equation*}
where \(\mathcal{L}_t\) denotes the training loss for task \(t\), \(\theta^{sh}\) denotes the shared model parameters, \(\theta^{i}\) denotes task-specific model parameters for task $i$, \(l_t\) denotes the empirical loss of task \(t\), and  $\lambda_{t}=\frac{1}{\bar{l}_{t} \sum_{i=1}^{T} \frac{1}{\bar{l}_{i}}}$. %Experiments show that the Tchebycheff loss outperforms traditional MTL objectives. 
The Tchebycheff loss can be combined with \added{aforementioned} adversarial MTL as \replaced{well}{in} \cite{lqh17}. %Tchebycheff loss dominates all previous MTL models.

Note that adjusting loss weight $\lambda_t$ of each task could guide the model to focus on different tasks during training while still learning multiple tasks at the same time, which can be seen as implicit task scheduling, compared to explicit task scheduling, which will be discussed in Section \ref{sec:task_scheduling}. In general, auxiliary MTL models are often bootstrapped with easier or lower-level tasks. For joint MTL, one would want to emphasize difficult tasks or tasks with lower homoscedasticity.
% Gradient Vaccine
\subsection{Gradient Regularization}
Aside from studying how to combine loss functions of different tasks, some studies optimize the training process by manipulating gradients. When jointly learning multiple tasks, the gradients from different tasks may be in conflict with each other, causing \deleted{negative}inter-task interference that harms performance.
\added{PCGrad \cite{ykglhf20} resolves such conflict using gradient projections.}
%As a remedy, PCGrad \cite{ykglhf20} directly projects conflicting gradients. 
Specifically, given two conflicting gradients $\mathbf{g}_i$ and $\mathbf{g}_j$ from tasks $i$ and $j$, respectively, PCGrad projects $\mathbf{g}_i$ onto the normal plane of $\mathbf{g}_j$ as
\begin{equation*}
\mathbf{g}_{i}^{\prime}=\mathbf{g}_{i}-\frac{\mathbf{g}_{i} \cdot \mathbf{g}_{j}}{\left\|\mathbf{g}_{j}\right\|^{2}} \mathbf{g}_{j}.
\end{equation*} Based on the observation that gradient similarity correlates well with language similarity and model performance, GradVac \cite{wtfc20}, which targets at optimization of \replaced{multilingual}{multi-lingual} models, regulates parameter updates according to geometry similarities between gradients. That is, GradVac alters both the direction and magnitude of gradients so that they are aligned with the cosine similarity between gradient vectors by modifying $\mathbf{g}_i$ as
\begin{equation*}
\mathbf{g}_{i}^{\prime}=\mathbf{g}_{i}+\frac{\left\|\mathbf{g}_{i}\right\|\left(\phi_{i j}^{T} \sqrt{1-\phi_{i j}^{2}}-\phi_{i j} \sqrt{1-\left(\phi_{i j}^{T}\right)^{2}}\right)}{\left\|\mathbf{g}_{j}\right\| \sqrt{1-\left(\phi_{i j}^{T}\right)^{2}}} \cdot \mathbf{g}_{j}
\end{equation*}
where $\phi_{i j} \in [-1,1]$ is the cosine distance between gradients $\mathbf{g}_i$ and $\mathbf{g}_j$. %The gradient similarity objective between task $i$ and task $j$ for the $k$-th layer $\phi_{ijk}^{(T)}$ is given by an exponential moving average over the course of training:
% \begin{equation}
% \hat{\phi}_{i j k}^{(t)}=(1-\beta) \hat{\phi}_{i j k}^{(t-1)}+\beta \phi_{i j k}^{(t)}
% \end{equation}
% where $\hat{\phi}_{ijk}^{(0)} = 0$ and $\beta$ is a hyper-parameter. 
Notice that PCGrad is a special case of GradVac when $\phi_{ij}^{T} = 0$. While PCGrad \replaced{does not modify positively associated gradients}{does nothing for gradients from positively associated tasks}, GradVac aligns both positively and negatively associated gradients, leading to a consist performance improvement for \replaced{multilingual}{multi-lingual} models.

\subsection{Data Sampling}
\label{sec:orge551827}

Machine learning models often suffer from imbalanced data distributions. MTL further complicates this issue in that training datasets of multiple tasks with potentially different sizes and data distributions are involved. \replaced{V}{To handle data imbalance, v}arious data sampling techniques have been proposed to properly construct training datasets. In practice, given \(M\) tasks and their datasets \(\{\mathcal{D}_1, \dots, \mathcal{D}_M\}\), a sampling weight \(p_t\) is assigned to task \(t\) to control the probability of sampling a data batch from \(\mathcal{D}_t\) in each training step.

% straightforward sampling strategy is proportional sampling \cite{swr19}, where the probability of sampling from a task is proportional to the size of its dataset as
%begin{equation}
%_t \propto |\mathcal{D}_t|.
%label{basic_sampling}
%end{equation}
%While adopted by many MTL models, proportional sampling (Eq. \eqref{basic_sampling}) favors tasks with larger datasets, thus increasing the risk of overfitting. To alleviate this problem, task-oriented sampling \cite{zxwj17} randomly samples the same amount of instances from all tasks, i.e.,
%\begin{equation*}
%p_t\propto\frac{1}{M}.
%\end{equation*}
%As a generalization of proportional sampling in Eq. (\ref{basic_sampling}), \(p_t\) for task \(t\) can take the following form as
In general, $p_t$ takes the form of:
\begin{equation*}
p_t \propto |\mathcal{D}_t|^\frac{1}{\alpha}
\end{equation*}
where ${\alpha}$ is the sampling temperature. When $\alpha > 1$, the divergence of sampling probabilities between tasks is reduced and vice versa. 
$\alpha$ can be \replaced{either}{viewed as} a \added{constant} hyperparameter\deleted{to be set by users} or \added{can} be changed dynamically during training. Similar to task loss weights, researchers have proposed various techniques to adjust \(\alpha\).
For example, the annealed sampling method \cite{sm19} adjusts \(\alpha\) as training proceeds. Given a total number of \(E\) epochs, \(\alpha\) at epoch \(e\) is set to
\begin{equation*}
\alpha(e) = \frac{1}{1 - \frac{0.8(e-1)}{E-1}}.
\end{equation*}
In this way, the model is trained more  evenly for different tasks towards the end of the training process to reduce inter-task interference. \added{\citet{wza20}} define \(\alpha\) as
\begin{equation*}
\alpha(e)=\min \left(\alpha_{m},(e-1) \frac{\alpha_{m}-\alpha_{0}}{M}+\alpha_{0}\right),
\end{equation*}
where \(\alpha_0\) and \(\alpha_m\) denote initial and maximum values of \(\alpha\)\deleted{, respectively}. The noise level of the self-supervised denoising autoencoding task is scheduled similarly, increasing difficulty after a warm-up period. In both works, temperature $\alpha$ increases during training which encourages up-sampling of low-resource tasks and alleviates overfitting.

\subsection{Task Scheduling}
\label{sec:task_scheduling}

Task scheduling determines the order of tasks \replaced{on}{in} which an MTL model is trained. A naive way is to train all tasks together. \deleted{For example, }\added{\citet{zxwj17}} take this way to train an MTL model, where data batches are organized as four-dimensional tensors of size \(N\times M\times T\times d\), where \(N\) denotes the number of samples, \(M\) denotes the number of tasks, \(T\) denotes sequence length, and \(d\) represents embedding dimensions. Similarly, \added{\citet{zh19}} put labeled data and unlabeled data together to form a batch and \added{\citet{xlz19}} learn the dependency parsing and semantic role labeling tasks together. In the case of auxiliary MTL, \added{\citet{as17}} train the primary task and one of the auxiliary tasks together at each step. Conversely, \added{\citet{sxzcy20}} train one of the primary tasks and the auxiliary task together and shuffles between the two primary tasks.

Alternatively, we can train an MTL model on different tasks at different steps. Similar to \deleted{the }data sampling techniques, we can assign a task sampling weight \(r_t\) for task \(t\), which is also called mixing ratio, to control the frequency of data batches from task \(t\).
The most common task scheduling technique is to shuffle between different tasks \cite{cw08,llsvk16,bs16,sg16,llzs16,pb17,stbp18,mthma18,gpb18b,scn18,mtdnd18,pcks18,ftl19,hlnd19,swr19,glzolzzdc19,tzwx19,jgkch20,rboc20}, either randomly or according to a pre-defined schedule.
While random shuffling is widely adopted, introducing more heuristics into scheduling could help further improving the performance of MTL models. For example, according to the similarity between each task and the primary task in a \replaced{multilingual}{multi-lingual} multi-task scenario, \added{\citet{lysj18}} define \(r_t\) as
\begin{equation*}
r_t = \mu_t\zeta_t|\mathcal{D}_t|^\frac{1}{2},
\end{equation*}
where \(\mu_t\) or \(\zeta_t\) is set to \(1\) if the corresponding task or language is the same as the primary task and $0.1$ otherwise.

Instead of using a fixed mixing ratio designed by hand, some researchers explore using a dynamic mixing ratio during the training process.
\added{\citet{gsa16}} schedule tasks by a state machine that switches between the two tasks and updates learning rate when validation loss rises.
\added{\citet{gpb18}} develop a  controller meta-network that dynamically schedules tasks based on multi-armed bandits. The controller has \(M\) arms and optimizes a control policy \(\pi_e\) for arm (task) \(t\) at step \(e\) based on an estimated action value \(Q_{e,t}\) defined as
\begin{equation*}
\begin{aligned}
\pi_e(t) &= \exp(Q_{e,t}/\tau)/\sum_{i=1}^M\exp(Q_{e,i}/\tau)\\
Q_{e,t} &= (1-\alpha)^eQ_{0,t} + \sum_{k=1}^e\alpha(1-\alpha)^{e-k}R_k
\end{aligned}
\end{equation*}
where $\tau$ denotes the temperature, \(\alpha\) is the decay rate, and \(R_k\) is the observed reward at step \(k\) that is defined as the negative validation loss of the primary task. Analysis shows that the bandit assigns a higher probability to the primary task at first and then \added{more evenly} switches between \added{all} tasks\deleted{ periodically}\added{, which echos the dynamic data sampling techniques introduced in Section \ref{sec:orge551827}}.
%Through experiments, the authors report that this kind of dynamic mixing ratio performs better than static mixing ratio.

Besides probabilistic approaches, task scheduling could also use heuristics based on certain performance metrics. By optimizing the Tchebycheff loss, \added{\citet{myld20}} learn from the task which has the worst validation performance at each step. The CA-MTL model \cite{php21} introduces an uncertainty-based sampling strategy based on Shannon entropy for joint learning of classification tasks. Specifically, given a batch size $b$ and $M$ tasks, a pool of $b\times M$ samples are first sampled. Then, the uncertainty measure $\mathcal{U}(x)$ for a sample $\mathbf{x}$ from task $i$ is defined as
\begin{equation*}
    \mathcal{U}\left(\mathbf{x}\right)=\frac{S_{i}\left(\mathbf{x}\right)}{\hat{S} \times S^{\prime}}
\end{equation*}
where $S$ denotes the Shannon entropy of the model's prediction on $\mathbf{x}$, $\hat S$ is the model's maximum average entropy over \added{the} $b$ samples from each task\replaced{.}{, and} $S^\prime$ denotes the entropy of a uniform distribution and is used to normalize the variance of the number of classes in each task. At last, $b$ samples with the highest uncertainty measures are used for training at the current step. Experiments show that this uncertainty-based sampling strategy could effectively avoid catastrophic forgetting and inter-task interference when jointly learning multiple tasks\replaced{, outperforming}{and outperforms} the aforementioned annealed sampling \cite{sm19}.
% \cite{myld20} optimizes the task that performs the worst on validation set at each step by optimizing a Tchebycheff loss.

In some cases, multiple tasks are learned sequentially. Such tasks usually form a clear dependency relationship or are of different difficulty levels. 
For instance, \added{\citet{ifmms17,nmktmo19}} train \deleted{their}MTL models on different tasks in the order of increasing difficulties.
%\cite{nmktmo19} focuses on learning easy tasks at the beginning and then shifts its focus to more difficult tasks through loss scheduling as described in Eq. (\ref{sigmoid_scheduling}).
Similarly, \added{\citet{hxts17}} train a multi-task model in the order of low-level tasks, high-level tasks, and at last mixed-level batches.
Unicoder \cite{hldgsjz19} trains its five pre-training objectives sequentially in each step.
\added{\citet{pvgr20}} first pre-train language and invertible adapters on language modeling before training task adapters on different down-stream tasks, where the language and invertible adapters can also receive gradient when training task adapters. To stabilize the training process when alternating between tasks with imbalanced dataset sizes, successive regularization \cite{hxts17,ftl19} can be added to loss functions as a regularization term, which is defined as
$\mathcal{L}_{sr} = \delta\left\|\theta_{e}-\theta_{e}^{\prime}\right\|^{2}$, where \(\theta_e\) and \(\theta'_e\) are model parameters before and after the update in the previous training step and \(\delta\) is a hyperparameter.

To sum up, task scheduling for MTL aims at alleviate overfitting and negative transfer caused by imbalanced dataset size. For auxiliary MTL, depending on the relationship between tasks, we can either start with the primary task before training primary and auxiliary tasks together or adopt a \textit{pre-train then fine-tune} approach \cite{lhh18,hlnd19,wclqlw20,cgzzwzs19}, which bootstraps the model with auxiliary tasks that are often easier or more data-rich. For joint MTL, we would like to choose tasks that are more likely to benefit the model. Generally, dynamic scheduling approaches like CA-MTL performs better than using a fixed mixing ratio. 

%For auxiliary MTL, some researchers adopt a \emph{pre-train then fine-tune} methodology, which trains auxiliary tasks first before fine-tuning on the down-stream primary tasks. 
%For example, \cite{lhh18} trains auxiliary POS tagging and domain prediction tasks first before the news headline popularity prediction task.
%\cite{hlnd19} trains document-level tasks first and then the aspect-level primary task by fixing document-level model parameters so that parameters for domain-level tasks are not affected by the small amount of aspect-level data.
%\cite{wclqlw20} first pre-trains on self-supervised tasks and then fine-tunes on the smaller disfluency detection data.
%Similarly, \cite{cgzzwzs19} first pre-trains on abundant question-answer pair data before fine-tuning on the low-resource question-review answer identification data.

\section{Application in NLP Tasks}
\label{sec:applications}

In this section, we summarize the application of multi-task learning in NLP tasks, including applying MTL to optimize certain primary tasks (i.e., Auxiliary MTL), to jointly learn multiple tasks (i.e., Joint MTL), and to improve the performance in \replaced{multilingual}{multi-lingual} multi-task and multimodal scenarios. Existing research works have also explored different ways to improve the performance and efficiency of MTL models, as well as using MTL to study the relatedness of different tasks. %We present similar tasks together and generally follow a chronological order.

\subsection{Auxiliary MTL}
\label{sec:app_auxMTL}

\begin{table}[htbp]
\caption{\label{auxTab}A summary of auxiliary MTL studies according to types of primary and auxiliary tasks involved. `W', `S', and `D' in the three rightmost columns represent word-level, sentence-level, and document-level tasks for auxiliary tasks, respectively. `LM' denotes language modeling tasks and `Gen' denotes text generation tasks. The `Architecture' column denotes the architecture used, where PFS denotes Parallel Feature Sharing, PFF denotes Parallel Feature Fusion, PMS denotes Parallel Multi-level Supervision, HP denotes Hierarchical Pipeline, and GAA denotes Generative Adversarial Architecture.}
\centering
\resizebox{\textwidth}{!}{
\begin{tabular}{|c|c|ccccc|c|c|c|}
\hline
Primary Task & Reference & \multicolumn{5}{|c|}{W} & S & D&Architecture\\
\hline
 &  & Tagging & Parsing & Chunking & LM & Gen & Classification & Classification&\\
\hline
 
 \multirow{11}{*}{\shortstack{Sequence\\ Tagging}} & \cite{as17} & \(\checkmark\) &  & \(\checkmark\) &  &  &  & &PFS\\   
 & \cite{cbyls20} &  &  &  &  &  & \(\checkmark\) & &PFS\\
 & \cite{ltn20} & \(\checkmark\) &  &  &  &  &  & &PFS\\
 & \cite{wclqlw20} & \(\checkmark\) &  &  &  &  & \(\checkmark\) & &PFS\\
 & \cite{ll17} & \(\checkmark\) &  &  &  &  & \(\checkmark\) & &PFF\\
 & \cite{rei17} &  &  &  & \(\checkmark\) &  &  & &PMS\\
 & \cite{wtnmt19} &  &  &  &  & \(\checkmark\) &  & &PMS\\
 & \cite{ifmms17} &  &  &  &  &  &  & \(\checkmark\)&HP\\
 & \cite{xlz19} &  & \(\checkmark\) &  &  &  &  & &HP\\
 & \cite{nnnosat19} &  &  &  &  & \(\checkmark\) &  & &HP\\
 & \cite{amd20} & \(\checkmark\) &  & \(\checkmark\) &  &  & &  & HP\\
\hline
 \multirow{15}{*}{Classification} & \cite{lhh18} & \(\checkmark\) &  &  &  &  &  & \(\checkmark\)&PFS\\
 & \cite{lwscdl18} &  &  &  &  &  &  & \(\checkmark\)&PFS\\
 & \cite{wrjns19} &  &  &  &  &  &  & \(\checkmark\)&PFF\\
 & \cite{klz18} &  &  &  &  &  & \(\checkmark\) & \(\checkmark\)&PFS\\
 & \cite{yesb19} &  &  &  &  &  &  & \(\checkmark\)&PFF\\
 & \cite{lzs19} &  &  &  &  &  &  & \(\checkmark\)&PFF\\
 & \cite{lc19} &  &  &  &  &  &  & \(\checkmark\)&PMS\\
 & \cite{mtdnd18} & \(\checkmark\) &  &  &  &  &  & &PMS\\
 & \cite{rly19} &  &  &  &  &  & \(\checkmark\) & &PMS\\
 & \cite{fy19} & \(\checkmark\) &  &  &  &  &  & &PMS\\
 & \cite{mxp19} &  &  &  &  &  & \(\checkmark\) & &HP\\
 & \cite{slf19} & \(\checkmark\) &  &  &  &  &  & &HP\\
 & \cite{yccwzs19} &  &  &  &  &  & \(\checkmark\) & &HP\\
 & \cite{sslf20} &  &  &  &  &  & \(\checkmark\) & \(\checkmark\)&HP\\
 & \cite{rcs20} &  &  &  &  &  &  & \(\checkmark\)&GAA\\
\hline
\multirow{11}{*}{\shortstack{Text \\Generation}} & \cite{dh17} &  &  &  & \(\checkmark\) &  &  &&PFS \\
 & \cite{llsvk16} &  & \(\checkmark\) &  &  & \(\checkmark\) &  & &PFS\\
 & \cite{wza20} &  &  &  & \(\checkmark\) & \(\checkmark\) &  && PFS\\
 & \cite{gpb18} &  &  &  &  & \(\checkmark\) &  & &PFS\\
 & \cite{gpb18b} &  &  &  &  & \(\checkmark\) &  & &PFS\\
 & \cite{sgbjcld19} & \(\checkmark\) &  &  &  & \(\checkmark\) & \(\checkmark\) & &PFS\\
 & \cite{zzh19} &  &  &  & \(\checkmark\) &  &  & &PFS\\
 & \cite{zbh18} & \(\checkmark\) & \(\checkmark\) &  &  &  &  & &PFF\\
 & \cite{clthhz20} & \(\checkmark\) &  &  &  &  &  & &PMS\\
 & \cite{zzw19} &  &  &  & \(\checkmark\) &  &  & &HP\\
 & \cite{rboc20} &  &  &  &  & \(\checkmark\) &  & &HP\\
\hline
\multirow{2}{*}{\shortstack{Representation\\ Learning}} & \cite{stbp18} &  & \(\checkmark\) &  & \(\checkmark\) &  & \(\checkmark\) & &PFS\\
 & \cite{wzlsz20} &  &  &  & \(\checkmark\) & \(\checkmark\) & \(\checkmark\) & \(\checkmark\)&PFS\\
 \hline
\end{tabular}
}
\end{table}

Auxiliary MTL aims to improve the performance of certain primary tasks by introducing auxiliary tasks and is widely used in the NLP field for different types of primary task\added{s}, \replaced{such as}{including} sequence tagging, classification, text generation, and representation learning. Table \ref{auxTab} \replaced{summarizes the}{generally show} types of auxiliary tasks used along with different types of primary tasks. As shown in Table \ref{auxTab}, auxiliary tasks are usually closely related to primary tasks.

Targeting sequence tagging tasks,
\added{\citet{rei17}} adds a language modeling objective into a sequence labeling model to counter the sparsity of named entities and make full use of training data.
\added{\citet{as17}} add five auxiliary tasks for scientific keyphrase boundary classification, including syntactic chunking, frame target annotation, hyperlink prediction, multi-word expression identification, and semantic super-sense tagging.
\added{\citet{ll17}} use opinion word extraction and sentence-level sentiment identification to assist aspect term extraction.
\added{\citet{ifmms17}} train an extractive summarization model together with an auxiliary document-level classification task.
\added{\citet{xzz18}} transfer knowledge from a large open-domain corpus to the data-scarce medical domain for Chinese word segmentation \replaced{using}{by developing} a parallel MTL architecture.
HanPaNE \cite{wtnmt19} improves NER for chemical compounds by jointly training a chemical compound paraphrase model.
\added{\citet{xlz19}} enhance Chinese semantic role labeling by adding a dependency parsing model and uses the output of dependency parsing as additional features.
\added{\citet{nnnosat19}} improve the evidence extraction capability of an explainable multi-hop QA model by viewing evidence extraction as an auxiliary summarization task.
\added{\citet{amd20}} improve \deleted{the }character-level diacritic restoration with word-level syntactic diacritization, POS tagging, and word segmentation.
In \cite{cbyls20}, the performance of argument mining is improved by the argument pairing task on review and rebuttal pairs of scientific papers.
\added{\citet{ltn20}} make use of the similarity between word sense disambiguation and metaphor detection to improve the performance of the latter task.
To handle the primary disfluency detection task, \added{\citet{wclqlw20}} pre-train two self-supervised tasks using constructed pseudo training data before fine-tuning on the primary task.

Researchers have also applied auxiliary MTL to classification tasks, such as explicit \cite{llzs16} and implicit \cite{lwwnw17} discourse relation classification.
To improve automatic rumor identification, \added{\citet{klz18}} jointly train on the stance classification and veracity prediction tasks.
\added{\citet{lhh18}} learn a headline popularity prediction model with the help of POS tagging and domain prediction.
\added{\citet{lzs19}} enhance a rumor detection model with user credibility features.
\added{\citet{fy19}} add a low-level grammatical role prediction task into a discourse coherence assessment model to help improve its performance.
\added{\citet{mxp19}} enhance the hashtag segmentation task by introducing an auxiliary task which predicts whether a given hashtag is single-token or multi-token.
In \cite{slf19}, text classification is boosted by learning the predominant sense of words.
\added{\citet{wrjns19}} assist the fake news detection task by stance classification.
\added{\citet{cgzzwzs19}} jointly learn the answer identification task with an auxiliary question answering task.
To improve slot filling performance for online shopping assistants, \added{\citet{glzolzzdc19}} add NER and segment tagging tasks as auxiliary tasks.
In \cite{sslf20}, the organization evaluation for student essays is learned together with the sentence and paragraph discourse element identification tasks.
\added{\citet{lc19}} model the stance detection task with the help of the sentiment classification and self-supervised stance lexicon tasks.
Generative adversarial MTL architectures are used to improve classification tasks as well. Targeting pharmacovigilance mining, \added{\citet{yesb19}} treat mining on different data sources as different tasks and applies self-supervised adversarial training as an auxiliary task to help the model combat the variation of data sources and produce more generalized features. Differently, \added{\citet{rcs20}} enhance a feature extractor through unsupervised adversarial training with a discriminator that is pre-trained with supervised data.
Sentiment classification models can be enhanced by POS tagging and gaze prediction \cite{mtdnd18}, label distribution learning \cite{zfszwy18}, unsupervised topic modeling \cite{wwsllszz20}, or domain adversarial training \cite{wfgwz18}. 
In \cite{wh16}, besides the shared base model, a separate model is built for each Microblog user as an auxiliary task.
\added{\citet{rly19}} estimate causality scores via Naranjo questionnaire, consisting of 10 multiple-choice questions, with sentence relevance classification as an auxiliary task.
\added{\citet{lwscdl18}} introduce an auxiliary task of selecting the passages containing the answers to assist a multi-answer question answering task.
\added{\citet{yccwzs19}} improve a community question answering model with an auxiliary question category classification task.
To counter data scarcity in the multi-choice question answering task, \added{\citet{jgkch20}} propose a multi-stage MTL model that is first coarsely pre-trained using a large out-of-domain natural language inference dataset and then fine-tuned on an in-domain dataset.

For text generation tasks, MTL is brought in to improve the quality of the generated text.
It is observed in \cite{dh17} that adding a target-side language modeling task on the decoder of a neural machine translation (NMT) model brings moderate but consistent performance gain.
\added{\citet{llsvk16}} learn a \replaced{multilingual}{multi-lingual} NMT model with constituency parsing and image caption generation as two auxiliary tasks.
Similarly, \added{\citet{zbh18}} learn an NMT model together with the help of NER, syntactic parsing, and semantic parsing tasks.
To make an NMT model aware of the vocabulary distribution of the retrieval corpus for query translation, \added{\citet{sba19}} add an unsupervised auxiliary task that learns continuous bag-of-words embeddings on the retrieval corpus in addition to the sentence-level parallel data. \deleted{Recently, }\added{\citet{wza20}} build a \replaced{multilingual}{multi-lingual} NMT system with source-side language modeling and target-side denoising autoencoder.
For the sentence simplification task, \added{\citet{gpb18}} use paraphrase generation and entailment generation as two auxiliary tasks.
\added{\citet{gpb18b}} build an abstractive summarization model with the question and entailment generation tasks as auxiliary tasks.
By improving a language modeling task through MTL, we can generate more natural and coherent text for question generation \cite{zzw19} or task-oriented dialogue generation \cite{zzh19}.
\added{\citet{sgbjcld19}} implement a semantic parser that jointly learns question type classification, entity mention detection, as well as a weakly supervised objective via question paraphrasing.
\added{\citet{clthhz20}} enhance a text-to-SQL semantic parser by adding explicit condition value detection and value-column mapping as auxiliary tasks.
\added{\citet{rboc20}} view hierarchical text classification, where each text may have several labels on different levels, as a generation task\deleted{s} by generating from more general labels to more specific ones, and an auxiliary task of generating in the opposite order is introduced to guide the model to treat high-level and low-level labels more equally and therefore learn more robust representations.

Besides tackling specific tasks, some researchers aim at building general-purpose text representations for future use in downstream tasks. For example, \added{\citet{stbp18}} learn sentence representations through multiple weakly related tasks, including learning skip-thought vectors, neural machine translation, constituency parsing, and natural language inference tasks. 
\added{\citet{wzlsz20}} train multi-role dialogue representations via unsupervised multi-task pre-training on reference prediction, word prediction, role prediction, and sentence generation.
As existing pre-trained models impose huge storage cost for the deployment, PinText \cite{zl19} learns user profile representations through learning custom word embeddings, which are obtained by minimizing the distance between positive engagement pairs based on user behaviors, including homefeed, related pins, and search queries, by sharing the embedding lookup table.

\subsection{Joint MTL}
\label{sec:orgf41ee94}

Different from auxiliary MTL, joint MTL models optimize its performance on several tasks simultaneously. Similar to auxiliary MTL, tasks in joint MTL are usually related to or complementary to each other. Table \ref{jointTab} gives an overview of task combinations used in joint MTL models.
In certain scenarios, we can even convert models following the traditional pipeline architecture as in single-task learning to joint MTL models so that different tasks can adapt to each other. For example, \added{\citet{pcks18}} convert the parsing of Alexa meaning representation language into three independent tagging tasks for intents, types, and properties, respectively. \added{\citet{sp19}} transform the pipeline relation between POS tagging and morphological tagging into a parallel relation and further builds a joint MTL model.

Joint MTL has been proven to be an effective way to improve the performance of standard NLP tasks. For instance, \added{\citet{hxts17}} train six tasks of different levels jointly, including POS tagging, chunking, dependency parsing, relatedness classification, and entailment classification.
\added{\citet{zxwj17}} apply parallel feature fusion to learn multiple classification tasks, including sentiment classification on movie and product reviews. Different from traditional pipeline methods, \added{\citet{lhoh18}} jointly learn identification and classification of entities, relations, and coreference clusters in scientific literatures.
\added{\citet{swr19}} optimize four semantic tasks together, including NER, entity mention detection (EMD), coreference resolution (CR), and relation extraction (RE) tasks.
\added{\citet{gsa16,zzl20,ylxscz19}} learn entity extraction alongside relation extraction.
For sentiment analysis tasks, \added{\citet{cjol18}} jointly learn dialogue act and sentiment recognition using the parallel feature sharing MTL architecture.
\added{\citet{hlnd19}} learn the aspect term extraction and aspect sentiment classification tasks jointly to facilitate aspect-based sentiment analysis.
\added{\citet{zhzlx20}} build a joint aspect term, opinion term, and aspect-opinion pair extraction model through MTL and shows that the joint model outperforms single-task and pipeline baselines by a large margin.

Besides well-studied NLP tasks, joint MTL is also widely applied in various downstream tasks. One major problem of such tasks is the lack of sufficient labeled data. Through joint MTL, one could take advantage of data-rich domains via implicit knowledge sharing. In addition, abundant unlabeled data could be utilized via unsupervised learning techniques.
\added{\citet{zlzw19}} develop a joint MTL model for the NER and entity name normalization tasks in the medical field.
\added{\citet{lhsflh18,zssxsl20}} use MTL to perform simile detection, which includes simile sentence classification and simile component extraction.
To analyze Twitter demographic data, \added{\citet{vvr17}} jointly learn classification models for genders, ages, political orientations, and locations.
The SLUICE network \cite{rbas19} is used to learn four different non-literal language detection tasks in English and German \cite{deg18}.
\added{\citet{nrc18}} jointly train a monolingual formality transfer model and a formality sensitive machine translation model between English and French.
For community question answering, \added{\citet{jmn18}} build an MTL model that extracts existing questions related to the current one and looks for question-comment threads that could answer the question at the same time.
To analyze the argumentative structure of scientific publications, \added{\citet{lgpe18}} optimize argumentative component identification, discourse role classification, citation context identification, subjective aspect classification, and summary relevance classification together with a dynamic weighting mechanism.
Considering the connection between sentence emotions and the use of the metaphor, \added{\citet{drls19}} jointly train a metaphor identification model with an emotion detection model.
To ensure the consistency between generated key phrases (short text) and headlines (long text), \added{\citet{nmktmo19}} train the two generative models jointly with a document category classification model and adds a hierarchical consistency loss based on the attention mechanism.
An MTL model is proposed in \cite{sxzcy20} to jointly perform zero pronoun detection, recovery, and resolution, and unlike \replaced{previous}{existing} works, it does not require external syntactic parsing tools.

\begin{table}[htbp]
\caption{\label{jointTab}A summary of joint MTL studies according to types of tasks involved. `W', `S', `D', and `O' in the four rightmost columns represent the word-level, sentence-level, and document-level tasks, and tasks of other abstract levels such as RE, respectively. A single checkmark could mean joint learning of multiple tasks of the same type. The `Architecture' column denotes the architecture used, where PFS denotes Parallel Feature Sharing, PFF denotes Parallel Feature Fusion, PMS denotes Parallel Multi-level Supervision, HFF denotes Hierarchical Feature Fusion, HP denotes Hierarchical Pipeline, and HIM denotes Hierarchical Interactive MTL.}
\centering
\begin{tabular}{|c|cc|c|c|c|c|}
\hline
Reference & \multicolumn{2}{|c|}{W}& S & D & O&Architecture\\
\hline
 & Tagging & Generation & Classification & Classification & Classification&\\
\hline
\cite{lhoh18} & \(\checkmark\) &  &  &  & \(\checkmark\)&PFS\\
\cite{deg18} & \(\checkmark\) &  &  &  & &PFS\\
\cite{nrc18} &  & \(\checkmark\) &  &  & &PFS\\
\cite{sxzcy20} & \(\checkmark\) &  & \(\checkmark\) &  & &PFS\\
\cite{gsa16} & \(\checkmark\) &  &  &  & \(\checkmark\)&PFS\\
\cite{ylxscz19} & \(\checkmark\) &  &  &  & \(\checkmark\)&PFS\\
\cite{gdsg20} & \(\checkmark\) &  &  &  & &PFS\\
\cite{cjol18} &  &  &  & \(\checkmark\) & &PFS\\
\cite{zhzlx20} & \(\checkmark\) &  &  &  & \(\checkmark\)&PFS\\
\cite{drls19} & \(\checkmark\) &  & \(\checkmark\) &  & &PFF\\
\cite{zxwj17} &  &  & \(\checkmark\) & \(\checkmark\) & &PFF\\
\cite{nmktmo19} &  & \(\checkmark\) &  & \(\checkmark\) & &PMS\\
\cite{pcks18} & \(\checkmark\) &  &  &  & &PMS\\
\cite{lgpe18} & \(\checkmark\) &  & \(\checkmark\) &  & &PMS\\
\cite{swr19} & \(\checkmark\) &  &  &  & \(\checkmark\)&PMS\\
\cite{lhsflh18} & \(\checkmark\) &  & \(\checkmark\) &  & &PMS\\
\cite{vvr17} &  &  &  & \(\checkmark\) & &HFF\\
\cite{hlnd19} & \(\checkmark\) &  & \(\checkmark\) &  & &HP\\
\cite{zlzw19} & \(\checkmark\) &  &  &  & &HP\\
\cite{zzl20} & \(\checkmark\) &  &  &  & \(\checkmark\)&HP\\
\cite{hxts17} & \(\checkmark\) &  & \(\checkmark\) & \(\checkmark\) & &HP\\
\cite{sp19} & \(\checkmark\) &  &  &  & &HP\\
\cite{zssxsl20} & \(\checkmark\) &  & \(\checkmark\) &  & &HIM\\
\hline
\end{tabular}
\end{table}

Moreover, joint MTL is suitable for multi-domain or multi-formalism NLP tasks. 
%Through MTL, the robustness and generalization ability of machine learning models are improved. 
Multi-domain tasks share the same problem definition and label space among tasks, but have different data distributions.
Applications in multi-domain NLP tasks include sentiment classification \cite{lz08,wh15}, dialog state tracking \cite{mstgsvwy15}, essay scoring \cite{czb16}, deceptive review detection \cite{hzcyll16}, multi-genre emotion detection and classification \cite{td18}, RST discourse parsing \cite{bps16}, historical spelling normalization \cite{bs16}, and document classification \cite{tzwx19}.
Multi-formalism tasks have the same problem definition but may have different while structurally similar label spaces.
\added{\citet{pts17,ks19}} model three different formalisms of semantic dependency parsing (i.e., DELPH-IN MRS (DM) \cite{fzk12a}, Predicate-Argument Structures (PAS) \cite{mkmmbfks94}, and Prague Semantic Dependencies (PSD) \cite{hhpsbcfmppo12}) jointly.
In \cite{har18}, a transition-based semantic parsing system is trained jointly on different parsing tasks, including Abstract Meaning Representation (AMR) \cite{bbcgghkkps13}, Semantic Dependency Parsing (SDP) \cite{okmzcfhiu16}, and Universal Dependencies (UD) \cite{ndgghmmppstz16}, and it shows that joint training improves performance on the testing UCCA dataset.
\added{\citet{llzs16}} jointly model discourse relation classification on two distinct datasets: PDTB and RST-DT.
\added{\citet{fov18}} show the dual annotation and joint learning of two distinct sets of relations for noun-noun compounds could improve the performance of both tasks.
In \cite{zh19}, an adversarial MTL model is proposed for morphological modeling for high-resource modern standard Arabic and its low-resource dialect Egyptian Arabic, to enable knowledge between the two domains. %The proposed model could effectively learn domain-invariant features and alleviate overfitting caused by the small and noisy dialect dataset.

\subsection{\replaced{Multilingual}{Multi-lingual} and Multimodal Tasks}
\label{sec:org5cdfd01}

\replaced{Multilingual}{Multi-lingual} machine learning has always been a hot topic in the NLP field with a representative example of NMT systems mentioned in Section \ref{sec:app_auxMTL}. Since monolingual data source may be limited and biased, leveraging data from multiple languages through MTL can benefit \replaced{multilingual}{multi-lingual} machine learning models, such as language intent learning in Japanese and English \cite{mthma18} and sentiment classification in Chinese and English \cite{wfgwz18}. Another use of MTL is cross-lingual knowledge transfer, where knowledge learned in one language can be used in tasks in another language. For example, \cite{nrc18} develops a formality-sensitive translation system from English to French where formality labels are only available in English. Besides, effort has also been made to learn unified cross-lingual language representations \cite{scn18,hldgsjz19}. Such cross-lingual representations could substantially boost performance under low-resource settings \cite{lysj18}.

One step further from \replaced{multilingual}{multi-lingual} learning, multimodal learning has attracted an increasing interest in recent years. Researchers have incorporated features from multiple modalities, such as auditory and visual features, to text-related cross-modal tasks. To this end, MTL is a natural choice for learning generalized multimodal features by shaping a shared cross-modal feature space. One example is end-to-end speech translation \cite{csll20} where speech recognition and text translation are learned jointly. Similarly for video captioning \cite{pb17}, the video prediction task and text entailment generation task are used to enhance the encoder and decoder of the model, respectively. A multimodal representation space also makes it possible to build natural language interfaces to different systems. One example is semantic navigation \cite{clspb20}, where an agent acts according to navigation commands in a 3-D environment. The key is learning a one-to-one mapping, also known as knowledge grounding, between visual feature maps and text tokens via joint learning of object detection and visual question answering tasks. A multi-task evaluation framework \cite{skvbefl20} is proposed to evaluate knowledge grounding of such vision-language models.

\subsection{Task Relatedness in MTL}
\label{sec:org6ffb1c6}

A key issue that affects the performance of MTL is how to properly choose a set of tasks for joint training. Generally, tasks that are similar and complementary to each other are suitable for multi-task learning, and there are some works that studies this issue for NLP tasks. For semantic sequence labeling tasks, \added{\citet{mp17}} report that MTL works best when the label distribution of auxiliary tasks has low kurtosis and high entropy. This finding also holds for rumor verification \cite{klz18}.
Similarly, \added{\citet{llzs16}} report that tasks with major differences, such as implicit and explicit discourse classification, may not benefit much from each other.
To quantitatively estimate the likelihood of two tasks benefiting from joint training, \added{\citet{sb20}} propose a dataset similarity metric which considers both tokens and their labels. The proposed metric is based on the normalized mutual information of the confusion matrix between label clusters of two datasets. Such similarity metrics could help identify helpful tasks and improve the performance of MTL models that are empirically hard to achieve through manual selection.

As MTL assumes certain relatedness and complementarity between the chosen tasks, the performance gain brought by MTL can in turn reveal the strength of such relatedness.
\added{\citet{chs18}} study the pairwise impact of joint training among 11 tasks under 3 different MTL schemes and show that MTL on a set of properly selected tasks outperforms MTL on all tasks. The harmful tasks either are totally unrelated to other tasks or possess a small dataset that \replaced{is prone to overfitting}{can be easily overfitted}. For dependency parsing problems, 
\added{\citet{pts17,ks19}} claim that MTL works best for formalisms that are more similar. 
\added{\citet{drls19}} model the interplay of the metaphor and emotion via MTL and reports that metaphorical features are beneficial to sentiment analysis tasks.
Unicoder \cite{hldgsjz19} presents results of jointly fine-tuning on different sets of languages as well as pairwise cross-language transfer among 15 languages, and finds that knowledge transfer between English, Spanish, and French is easier than other \replaced{combinations}{sets} of languages.

\section{Data Source and Benchmarks for Multi-task Learning}
\label{sec:data}

In this section, we introduce the ways of preparing datasets for training MTL models and some benchmark datasets.

\subsection{Data Source}
\label{sec:orgbec0c29}

Given \(M\) tasks with corresponding datasets \(\mathcal{D}_t = \{\mathbf{X}_t, \mathbf{Y}_t\}, t=1,\ldots, M\), where \(\mathbf{X}_t\) denotes the set of data instances in task $t$ and \(\mathbf{Y}_t\) denotes the corresponding labels, we denote the entire dataset for the $M$ tasks by \(\mathcal{D} = \{\mathbf{X},\mathbf{Y}\}\). We describe different forms of \(\mathcal{D}\) in the following sections.

\subsubsection{Disjoint Datasets}
\label{sec:orgbedfbd4}

In most multi-task learning literature, the \replaced{datasets}{training sets} of different tasks have distinct label spaces, i.e. \(\forall i\neq j,\; \mathbf{Y}_i \cap \mathbf{Y}_j=\emptyset\). In this case, \(\mathcal{D} = \{\mathcal{D}_1 ,\dots,\mathcal{D}_M\}\)\added{.}
The most popular way to train MTL models on such tasks is to alternate between different tasks \cite{cw08,llsvk16,bs16,sg16,gsa16,lqh16,llzs16,pb17,dh17,hxts17,mthma18,xzcwj18b,zcq18,ftl19,rly19}, either randomly or by a \replaced{schedule,}{pre-defined order.} \replaced{as previously discussed in}{Thus the model handles data from different datasets in turns as discussed in} Section \ref{sec:optimization}.

\subsubsection{Multi-label Datasets}
\label{sec:org9fa879e}
Instances in multi-label datasets \replaced{share one feature space for all tasks}{have one label space for each task}, i.e. \(\forall i\neq j,\;\mathbf{X}= \mathbf{X}_i=\mathbf{X}_j\), which makes it possible to optimize all task-specific components at the same time. In this case, \(\mathcal{D} = \{\mathbf{X}, \hat{ \mathbf{Y}}\}\) where \(\hat{\mathbf{Y}}=\cup_{i=1}^M\mathbf{Y}_i\).

Multi-label datasets can be created by giving extra \deleted{manual}annotations to existing data. For example,
\added{\citet{pts17,ks19}} annotate dependency parse trees of three different formalisms for each text input.
\added{\citet{vvr17}} label Twitter posts with 4 demographic labels.
\added{\citet{fov18}} annotate two distinct sets of relations over the same set of underlying chemical compounds.

The extra annotations can be created automatically as well, resulting in a self-supervised multi-label dataset. Extra labels can be obtained using pre-defined rules \cite{rei17,lc19}.
In \cite{lwwnw17}, to synthesize unlabeled dataset for the auxiliary unsupervised implicit discourse classification task, explicit discourse connectives (e.g., because, but, etc.) are removed from a large corpus and \deleted{such connectives are }used as implicit relation labels.
\added{\citet{nrc18}} combine an English corpus with formality labels and an unlabeled English-French parallel corpus by random selection and concatenation to facilitate the joint training of formality style transfer and formality-sensitive translation.
\added{\citet{td18}} use hashtags to represent genres of tweet posts.
\added{\citet{wtnmt19}} generate sentence pairs by replacing chemical named entities with their paraphrases in the PubChemDic database.
Unicoder \cite{hldgsjz19} uses translated text from the source language to fine-tune on the target language.
\added{\citet{wclqlw20}} create disfluent sentences by randomly repeating or inserting $n$-grams.
Besides annotating in the aforementioned ways, some researchers create self-supervised labels with the help of external tools or previously trained models.
\added{\citet{slf19}} obtain dominant word sense labels from WordNet \cite{f10}.
\added{\citet{dxlydfls19}} apply entity linking for QA data over databases through an entity linker.
\added{\citet{glzolzzdc19}} assign NER and segmentation labels for three tasks using an unsupervised dynamic programming method.
\added{\citet{llcp20}} use the output of a meta-network as labels for unsupervised training data.
As a special case of multi-label dataset, mask orchestration \cite{wzlsz20} provides different parts of an instance to different tasks by applying different masks. That is, labels for one task may become the input for another task\deleted{and vice versa}.

\subsection{Multi-task Benchmark Datasets}
\label{sec:orgcc85f7a}

\begin{table}[!htbp]
\caption{\label{datasetTab}Statistics of multi-task benchmark datasets for NLP tasks.}
\centering
% \resizebox{\textwidth}{!}{
\begin{tabular}{|c|c|c|c|c|}
\hline
Dataset & \# Tasks & \# Languages & \# Samples & Topic \\ 
\hline
 GLUE \cite{wsmhlb19} & 9 & 1 (en)& 2157k & Language Understanding \\
 Super GLUE \cite{wpnsmhlb19} & 8 &1 (en)& 160k & Language Understanding \\
 MMMLU \cite{hbbzmss20} & 57 &  1 (en) & - & Language Understanding \\ 
 Xtreme \cite{hrsnfj20} & 9 & 40 & 597k & \replaced{Multilingual}{Multi-lingual} Learning \\
 XGLUE \cite{ldgwgqgsjcfzacwbqcwlycmz20} & 11 & 100 & 2747G & Cross-lingual Pre-training \\
 LSParD \cite{sgbjcld19} & 3 & 1 (en) & 51k & Semantic Parsing \\
 ECSA \cite{glzolzzdc19} & 3 & 1 (cn) & 28k & Language Processing \\
 ABC \cite{gbhws20} & 4 & 1 (en) & 5k & Anti-reflexive Gender Bias Detection \\
 CompGuessWhat?! \cite{skvbefl20} & 4 & 1 (en) & 66k & Grounded Language Learning \\
 SCIERC \cite{lhoh18} & 3 & 1 (en) & 500 & Scientific Literature Understanding \\
 \hline
\end{tabular}
% }
\end{table}

As summarized in Table \ref{datasetTab}, we list a few public multi-task benchmark datasets for NLP tasks. 
\begin{itemize}
\item \textbf{GLUE} \cite{wsmhlb19} is a benchmark dataset for evaluating natural language understanding (NLU) models. The main benchmark consists of 8 sentence and sentence-pair classification tasks as well as a regression task. The tasks cover a diverse range of genres, dataset sizes, and difficulties. Besides, a diagnostic dataset is provided to evaluate the ability of NLU models on capturing a pre-defined set of language phenomena.

\item \textbf{SuperGLUE} \cite{wpnsmhlb19} is a generalization of GLUE. As the performance of state-of-the-art models has exceeded non-expert human baselines on GLUE, SuperGLUE contains a set of 8 more challenging NLU tasks along with comprehensive human baselines. Besides retaining the two hardest tasks in GLUE, 6 tasks are added with two new question formats: coreference resolution and question answering (QA).

\item \textbf{Measuring Massive Multitask Language Understanding (MMMLU)} \cite{hbbzmss20} is a multi-task few-shot learning dataset for world knowledge and problem solving abilities of language processing models. This dataset covers 57 subjects including 19 in STEM, 13 in humanities, 12 in social sciences, and 13 in other subjects. This dataset is split into a few-shot development set that has 5 questions for each subject, a validation set for tuning hyper-parameters containing 1540 questions, and a test set with 14079 questions.

\item \textbf{Xtreme} \cite{hrsnfj20} is a multi-task benchmark dataset for evaluating cross-lingual generalization capabilities of \replaced{multilingual}{multi-lingual} representations covering 9 tasks in 40 languages. The tasks include 2 classification tasks, 2 structure prediction tasks, 3 question answering tasks, and 2 sentence retrieval tasks. Out of the 40 languages involved, 19 languages appear in at least 3 datasets and the rest 21 languages appear in at least one dataset.% and are form less represented language families.

\item \textbf{XGLUE} \cite{ldgwgqgsjcfzacwbqcwlycmz20} is a benchmark dataset that supports the development and evaluation of large cross-lingual pre-trained language models. The XGLUE dataset includes 11 downstream tasks, including 3 single-input understanding tasks, 6 pair-input understanding tasks, and 2 generation tasks. The pre-training corpus consists of a small corpus that includes a 101G \replaced{multilingual}{multi-lingual} corpus covering 100 languages and a 146G bilingual corpus covering 27 languages, and a large corpus with 2,500G \replaced{multilingual}{multi-lingual} data covering 89 languages.

\item \textbf{LSParD} \cite{sgbjcld19} is a multi-task semantic parsing dataset with 3 tasks, including question type classification, entity mention detection, and question semantic parsing. Each logical form is associated with a question and multiple human annotated paraphrases. This dataset contains 51,164 questions in 9 categories, 3361 logical form patterns, and 23,144 entities.

\item \textbf{ECSA} \cite{glzolzzdc19} is a dataset for slot filling, named entity recognition, and segmentation to evaluate online shopping assistant systems in Chinese. The training part contains 24,892 pairs of input utterances and their corresponding slot labels, named entity labels, and segment labels. The testing part includes 2,723 such pairs with an Out-of-Vocabulary (OOV) rate of 85.3\%, which is much higher than the ATIS dataset \cite{hgd90} whose OOV \added{rate} is smaller than 1\%.

\item \textbf{ABC} \cite{gbhws20}, the Anti-reflexive Bias Challenge, is a multi-task benchmark dataset designed for evaluating gender assumptions in NLP models. ABC consists of 4 tasks, including language modeling, natural language inference (NLI), coreference resolution, and machine translation. A total of 4,560 samples are collected by a template-based method. The language modeling task is to predict the pronoun of a sentence. For NLI and coreference resolution, three variations of each sentence are used to construct entailment pairs. For machine translation, sentences with two variations of third-person pronouns in English are used as source sentences.
\item \textbf{CompGuessWhat?!} \cite{skvbefl20} is a dataset for grounded language learning with 65,700 collected dialogues. It is an instance of the Grounded Language Learning with Attributes (GROLLA) framework. The evaluation process includes three parts: goal-oriented evaluation (e.g., Visual QA and Visual NLI), object attribute prediction, and zero-shot evaluation. %The authors hypothesize that good grounded representations should be expressive enough to capture the attribute of target objects besides performing the intended goal-oriented task.

\item \textbf{SCIERC} \cite{lhoh18} is a multi-label dataset for identifying entities, relations, and cross-sentence coreference clusters from abstracts of research papers. %, supporting the development of unified models that jointly solve these tasks. 
SCIERC contains 500 scientific abstracts collected from proceedings in 12 conferences and workshops in artificial intelligence.
\end{itemize}

\section{Conclusion and Discussions}
\label{sec:conclusion}

In this paper, we give an overview of the application of multi-task learning in recent natural language processing research, focusing on deep learning approaches. We first present different architectures of MTL used in recent research literature, including parallel architecture, hierarchical architecture, modular architecture, and generative adversarial architectures. After that, optimization techniques, including loss construction, data sampling, and task scheduling are discussed. After briefly summarizing the application of MTL in different down-stream tasks, we describe the ways to manage data sources in MTL as well as some MTL benchmark datasets for NLP research.

There are several directions worth further investigations for future studies. Firstly, given multiple NLP tasks, how to find a set of tasks that could take advantage of MTL remains a challenge. Besides improving performance of MTL models, a deeper understanding of task relatedness could also help expanding the application of MTL to more tasks. Though there are some works studying this issue, as discussed in Section \ref{sec:org6ffb1c6}, they are far from mature.
%This issue is important since studying it can not only help understand the rationale that why MTL can help improve the performance but also broaden the use of MTL in the NLP field. We can do further studies on this issue.

Secondly, current NLP models often rely on a large or even huge amount of labeled data. However, in many real-world applications, where large-scale data annotation is costly, this requirement cannot be easily satisfied. In this case, we may consider to leverage abundant unlabeled data in MTL by using self-supervised or unsupervised learning techniques.
%we may need to combine MTL with unsupervised or self-supervised learning techniques to explore the potential information contained in abundant unlabeled data to improve the performance of all the tasks.

Thirdly, we are curious about whether we can create more powerful Pre-trained Language Models (PLMs) via more advanced MTL techniques. PLMs have become an essential part of NLP pipeline. Though most PLMs are trained on multiple tasks, the MTL architectures used are mostly simple feature sharing architectures\added{.} A better MTL architecture might be the key for the next breakthrough for PLMs. 

At last, it would be interesting to extend the use of MTL to more NLP tasks. Though there are many NLP tasks that can be jointly learned by MTL, most NLP tasks are well-studied tasks, such as classification, sequence labeling, and text generation, as shown in Tables \ref{auxTab} and \ref{jointTab}. We would like to see how MTL could benefit more challenging NLP tasks, such as building dialogue systems and multi-modal learning tasks. % This paragraph could be improved.

\section*{Acknowledgements}

This work is supported by NSFC key grant under grant no. 62136005, NSFC general grant under grant no. 62076118, and Shenzhen fundamental research program JCYJ20210324105000003.

\bibliography{MTL_APP_NLP}
\bibliographystyle{ACM-Reference-Format}
\end{document}